\documentclass[conference]{IEEEtran}
\IEEEoverridecommandlockouts
\usepackage{cite}
\usepackage{amsmath,amssymb,amsfonts}
\usepackage{algorithmic}
\usepackage{graphicx}
\usepackage{textcomp}
\usepackage{xcolor}

\usepackage{caption}
\usepackage{subcaption} 
\usepackage{booktabs} 
\usepackage{array}
\usepackage{multicol}
\usepackage{colortbl}
\usepackage{tabularx}
\usepackage{multirow}
\usepackage{hyperref}

\definecolor{lightpink}{RGB}{255,229,229}       
\definecolor{pink}{RGB}{255,193,193}            
\definecolor{red}{RGB}{255,0,0}                 
\definecolor{darkred}{RGB}{204,0,0}             
\definecolor{lightgreen1}{RGB}{204,255,204}     
\definecolor{lightgreen2}{RGB}{153,255,153}     
\definecolor{lightgreen3}{RGB}{102,255,102}     
\definecolor{lightblue1}{RGB}{229,255,229}      
\definecolor{lightgray}{RGB}{229,229,229}       
\definecolor{lightbrown1}{RGB}{255,229,180}     
\definecolor{lightyellow1}{RGB}{255,255,229}    
\definecolor{lightyellow2}{RGB}{255,255,180}    
\definecolor{brown1}{RGB}{255,204,153}          
\definecolor{lightblue2}{RGB}{204,229,255}      
\definecolor{blue}{RGB}{153,204,255}            
\definecolor{targetred}{RGB}{255,0,0}           
\definecolor{targetgreen}{RGB}{0,100,0}         
\definecolor{targetyellow}{RGB}{255,255,204}    
\definecolor{targetblue}{RGB}{100,149,237}      

\def\BibTeX{{\rm B\kern-.05em{\sc i\kern-.025em b}\kern-.08em
    T\kern-.1667em\lower.7ex\hbox{E}\kern-.125emX}}
\begin{document}

\title{A Dual-Branch Local-Global Framework for Cross-Resolution Land Cover Mapping}

\author{

\IEEEauthorblockN{
Peng Gao\textsuperscript{1,2},
Ke Li\textsuperscript{2},
Di Wang\textsuperscript{1,2},
Yongshan Zhu\textsuperscript{2},
Yiming Zhang\textsuperscript{3},
Xuemei Luo\textsuperscript{2},
Yifeng Wang\textsuperscript{2}
}

\IEEEauthorblockA{
\textsuperscript{1}\textit{Guangzhou Institute of Technology}, \textit{Xidian University}, Guangzhou, China\\
\textsuperscript{2}\textit{School of Computer Science and Technology}, \textit{Xidian University}, Xi'an, China\\
\textsuperscript{3}\textit{Department of Mathematics}, \textit{University of California}, San Diego, USA\\
}
}

\maketitle

\begin{abstract}
Cross-resolution land cover mapping aims to produce high-resolution semantic predictions from coarse or low-resolution supervision, yet the severe resolution mismatch makes effective learning highly challenging.
Existing weakly supervised approaches often struggle to align fine-grained spatial structures with coarse labels, leading to noisy supervision and degraded mapping accuracy.
To tackle this problem, we propose DDTM, a dual-branch weakly supervised framework that explicitly decouples local semantic refinement from global contextual reasoning.
Specifically, DDTM introduces a diffusion-based branch to progressively refine fine-scale local semantics under coarse supervision, while a transformer-based branch enforces long-range contextual consistency across large spatial extents.
In addition, we design a pseudo-label confidence evaluation module to mitigate noise induced by cross-resolution inconsistencies and to selectively exploit reliable supervisory signals.
Extensive experiments demonstrate that DDTM establishes a new state-of-the-art on the Chesapeake Bay benchmark, achieving 66.52\% mIoU and substantially outperforming prior weakly supervised methods.
The code is available at \url{https://github.com/gpgpgp123/DDTM}.
\end{abstract}

\begin{IEEEkeywords}
Cross-resolution land cover mapping, weakly supervised learning, diffusion models, transformer, semantic segmentation.
\end{IEEEkeywords}

\begin{figure*}[ht]
    \centering
    \centerline{\includegraphics[width=\linewidth]{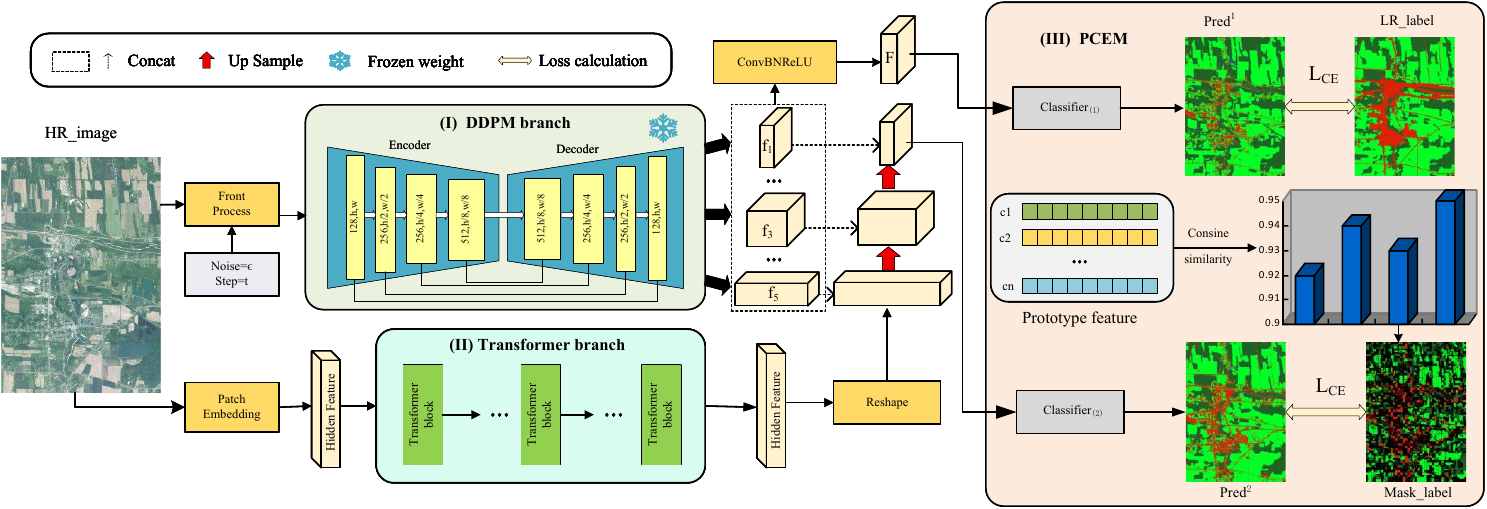}}
    \caption{Architecture overview of the proposed DDTM.}
    \label{fig:1}
\end{figure*}

\section{Introduction}
\label{sec:intro}
Recent advances in satellite remote sensing have enabled the acquisition of abundant high-resolution imagery, revealing fine-grained spatial structures such as textures, boundaries, and object shapes across large geographic regions \cite{cihlar2000land}.
These developments have made high-resolution land cover mapping an increasingly important dense prediction task, aiming to assign a semantic category to each pixel, which is critical for environmental monitoring, urban planning, and resource management \cite{li2025segearth,li2025rsvg,li2024language,li2023mixing}.
Despite the availability of rich visual information, constructing large-scale high-resolution land cover maps remains challenging due to the prohibitive cost of pixel-level annotation.

To overcome this limitation, recent studies have explored weakly supervised learning \cite{cao2022coarse} using coarse or historical land cover products (LCPs) \cite{chen2015global, gong2019stable, zanaga2022esa} as supervision.
However, weak supervision in land cover mapping inherently suffers from severe cross-resolution mismatch: fine-grained spatial structures in high-resolution imagery must be learned from low-resolution and error-prone labels.
This mismatch introduces substantial ambiguity in supervision, particularly around object boundaries and heterogeneous regions, which often leads to noisy pseudo labels and degraded mapping accuracy.

Existing weakly supervised methods predominantly rely on convolutional neural networks (CNNs) \cite{li2022breaking}, which are effective for local pattern extraction but limited in modeling complex fine-scale semantics under ambiguous supervision.
Moreover, CNNs struggle to explicitly enforce global semantic consistency over large spatial extents, which is crucial when supervision lacks spatial precision.
As a result, current approaches often fail to reconcile fine-grained local details with coarse global annotations in cross-resolution land cover mapping.

In this work, we revisit weakly supervised land cover mapping from a cross-resolution perspective and ask a fundamental question: \emph{how can fine-grained local semantics be reliably recovered while maintaining global semantic consistency under coarse supervision?}
To address this challenge, we observe two complementary inductive priors.
\textit{(i)} Denoising diffusion probabilistic models (DDPMs) have demonstrated strong capability in progressively refining high-dimensional data distributions, making them well suited for recovering fine-scale spatial structures from coarse supervision \cite{ho2020denoising, baranchuk2021label}.
\textit{(ii)} Transformer architectures excel at modeling long-range contextual dependencies, enabling global reasoning and semantic alignment across large spatial regions \cite{li2023mixing,li2024diagswin}.

Motivated by these insights, we propose DDTM, a local-global dual-branch framework for weakly supervised cross-resolution land cover mapping.
DDTM explicitly decouples local semantic refinement and global contextual reasoning through a diffusion-based branch and a transformer-based branch, respectively.
Furthermore, we introduce a pseudo-label confidence evaluation module to mitigate noise caused by resolution mismatch by selectively exploiting reliable supervisory signals from coarse annotations.
Together, these components enable robust feature refinement and significantly improve high-resolution land cover mapping performance under weak supervision.

Our contributions are summarized as follows:
\begin{itemize}
\item We study weakly supervised land cover mapping under cross-resolution supervision and highlight the key challenge of reconciling fine-grained structures with coarse and noisy labels.
\item We propose DDTM, a local--global dual-branch framework that decouples local semantic refinement and global contextual reasoning via a diffusion-based branch and a transformer-based branch.
\item We introduce a pseudo-label confidence evaluation module to alleviate supervision noise arising from resolution mismatch by selectively leveraging reliable supervisory signals.
\item Extensive experiments on the Chesapeake Bay benchmark demonstrate that DDTM achieves state-of-the-art performance, and ablation studies validate the effectiveness of each component.
\end{itemize}

\section{Related Work}
\label{sec:rw}
\subsection{Land Cover Mapping}
Land cover mapping has been extensively studied in remote sensing and computer vision, evolving from traditional pixel-wise classifiers to modern deep learning-based dense prediction models.
Early approaches relied on handcrafted features combined with classifiers such as Random Forests \cite{chan2008evaluation} and Support Vector Machines \cite{shi2015support}, which often failed to incorporate spatial context and resulted in noisy predictions in high-resolution imagery.
The introduction of CNNs, exemplified by U-Net \cite{ronneberger2015u}, enabled end-to-end pixel-level prediction by leveraging hierarchical feature representations.
To overcome the limited receptive fields of CNNs, Transformer-based architectures have recently been explored for land cover mapping \cite{vaswani2017attention, liu2021swin}.
By modeling long-range dependencies through self-attention, these methods improve global contextual understanding and have demonstrated promising performance in complex scenes.
However, most existing land cover mapping methods rely on fully supervised learning, where performance is tightly coupled with the availability of large-scale, high-quality pixel-level annotations.
The high cost of manual labeling remains a major bottleneck for scaling high-resolution land cover mapping to large geographic regions.

\subsection{Weakly Supervised Land Cover Mapping}
To reduce annotation cost, weakly supervised land cover mapping has attracted increasing attention.
These methods leverage coarse, sparse, or noisy supervision, such as low-resolution maps or historical products, to guide high-resolution prediction.
A representative approach is L2HNet \cite{li2022breaking}, which adopts a low-to-high framework to transfer supervision from coarse labels to high-resolution feature representations.
Despite its effectiveness, L2HNet is sensitive to noise induced by resolution mismatch, which often leads to inaccurate localization and degraded boundary quality.
More recently, Transformer-based weakly supervised methods, such as Paraformer \cite{li2024learning}, have been proposed to enhance global contextual modeling.
By employing parallel transformer structures, these methods improve large-scale semantic consistency under weak supervision.
Nevertheless, they typically depend on pseudo-label generation and self-training, making them vulnerable to accumulated label noise.
Moreover, existing weakly supervised approaches generally lack explicit mechanisms to jointly address fine-grained local refinement and global contextual reasoning under cross-resolution supervision, which limits their ability to recover detailed structures while maintaining semantic consistency.

In contrast, our work explicitly decouples local semantic refinement and global contextual modeling in a unified framework, enabling more robust learning from coarse supervision for high-resolution land cover mapping.

\section{Proposed Method}
\label{sec:method}
\subsection{Overview}
We propose DDTM, a dual-branch framework for weakly supervised cross-resolution land cover mapping, as illustrated in Fig.~\ref{fig:1}, which explicitly decouples local semantic refinement and global contextual reasoning.
Specifically, the diffusion branch focuses on recovering fine-grained local structures from noisy supervision, while the transformer branch enforces long-range semantic consistency.
A confidence-aware pseudo-labeling strategy is further introduced to mitigate supervision noise caused by resolution mismatch.

\subsection{Diffusion-based Local Semantic Refinement}
Weak supervision from coarse land cover products provides limited spatial precision, making it challenging to recover fine-scale structures such as boundaries and heterogeneous regions.
To address this issue, we leverage DDPMs as a powerful prior for progressive local semantic refinement.

DDPM consists of a forward noising process and a learned reverse denoising process.
To train the denoising network $f_{\theta}$, we leverage large-scale open-source remote sensing image datasets.
During training, each input image $x$ is progressively corrupted through the forward diffusion process to obtain a noisy sample $x_t$, where the noise level is controlled by the diffusion step $t$.
Specifically, Gaussian noise $\epsilon \sim \mathcal{N}(0, I)$ is added to the image, and the network $f_{\theta}$ is optimized to predict the injected noise using a mean squared error (MSE) loss:
\begin{equation}
\mathcal{L}_{\text{diff}} = \left\| f_{\theta}(x_t, t) - \epsilon \right\|_2^2, \quad \epsilon \sim \mathcal{N}(0, I).
\label{eq_1}
\end{equation}
Instead of using DDPM for image generation, we exploit a pretrained denoising network $f_{\theta}$ as a multi-scale feature extractor within our dual-branch framework.
The weights of $f_{\theta}$ are frozen during downstream training to avoid overfitting to noisy supervision and to provide a stable local semantic prior.
By adjusting the diffusion step, the model produces features corresponding to different noise levels, which naturally encode multi-scale semantic information.

Formally, given an input image $x$, the decoder of the pretrained DDPM outputs a set of intermediate feature maps:
\begin{equation}
[f_1, f_2, f_3, f_4, f_5] = f_{\theta}(x, x_{\text{step}}, \text{step}),
\label{eq_2}
\end{equation}
where $\text{step}$ denotes the diffusion step controlling noise intensity, and $f_i$ represents the decoder feature at the $i$-th scale.
Each feature map $f_i$ is upsampled to the original image resolution to obtain $f_i'$.
All upsampled features are then concatenated and passed through a ConvBR block, consisting of a convolution layer, batch normalization, and ReLU activation, to perform channel adjustment and normalization:
\begin{equation}
F = \text{ConvBR}(\text{Cat}([f_i'])),\ \text{where}\ i=\{1,\dots,5\}.
\label{eq_3}
\end{equation}
Benefiting from the progressive denoising process, the extracted features preserve fine-grained spatial textures and boundary details that are difficult to recover using conventional CNN-based encoders, which is particularly beneficial for cross-resolution land cover mapping under weak supervision.

\subsection{Transformer-based Global Context Modeling}
While the diffusion branch focuses on refining fine-grained local semantics, it lacks explicit mechanisms for modeling long-range dependencies across large spatial extents.
Such global contextual reasoning is crucial in cross-resolution land cover mapping, where coarse supervision provides only weak spatial guidance.
To complement this limitation, we introduce a Transformer branch to enforce global semantic consistency.

The Transformer backbone consists of 12 standard self-attention blocks, each composed of layer normalization, multi-head self-attention, and a feed-forward MLP.
Given an input image $x$, patch embeddings are first computed and augmented with positional encodings, producing a sequence of hidden representations.
Global dependencies are then modeled through multi-head self-attention:
\begin{equation}
\text{Attention}(Q, K, V) = \text{Softmax}\left( \frac{QK^\top}{\sqrt{d_k}} \right)V,
\label{eq:attn}
\end{equation}
where $Q$, $K$, and $V$ denote the query, key, and value projections, respectively.

The encoded Transformer features are progressively upsampled to the high-resolution spatial scale.
At each scale, they are fused with the corresponding multi-scale features extracted by the diffusion branch.
Specifically, the Transformer features and diffusion features are concatenated and passed through a ConvBR block for channel adjustment and normalization.
This hierarchical fusion enables the model to jointly leverage fine-grained local semantics from the diffusion branch and long-range contextual information from the Transformer branch, yielding the final feature representation $F$ for prediction.

\begin{table*}[!t]
    \centering
    \caption{Quantitative comparison (mIoU) on the Chesapeake Bay dataset across six states.}
    \label{tab:partial_miou_comparison}
    \resizebox{\linewidth}{!}{
    \begin{tabular}{lcccccccc}
        \toprule
        Methods & Venue & Delaware & New York & Maryland & Pennsylvania & Virginia & West Virginia & Average \\
        \midrule
        SkipFCN \cite{li2021change} & IGARSS'21 & 60.97 & 64.83 & 59.44 & 55.37 & 64.72 & 54.66 & 60.00 \\
        DC-Swin \cite{wang2022novel} & GRSL'22 & 59.65 & 65.99 & 58.60 & 58.06 & 64.11 & 48.15 & 59.09 \\
        UNetFormer \cite{wang2022unetformer} & ISPRS'22 & 58.85 & 65.11 & 61.34 & 59.10 & 60.84 & 47.20 & 58.74 \\
        EfficientViT \cite{cai2023efficientvit} & ICCV'23 & 53.72 & 61.28 & 59.48 & 51.38 & 57.34 & 48.76 & 55.33 \\
        MobileViT \cite{mehta2021mobilevit} & ICLR'22 & 58.03 & 61.32 & 61.84 & 55.53 & 57.04 & 48.64 & 57.07 \\
        CoAtNet \cite{dai2021coatnet} & NeurIPS'21 & 56.89 & 62.83 & 61.25 & 53.57 & 65.67 & 51.34 & 58.59 \\
        ConViT \cite{d2021convit} & ICML'21 & 55.26 & 60.71 & 61.58 & 53.94 & 59.80 & 49.11 & 56.73 \\
        L2HNet \cite{li2022breaking} & ISPRS'22 & 61.77 & 68.12 & 65.24 & 58.52 & \underline{69.39} & \underline{55.43} & 63.08 \\
        TransUNet \cite{chen2024transunet} & MIA'24 & 53.15 & 60.53 & 60.42 & 51.08 & 66.21 & 47.52 & 56.49 \\
        Paraformer \cite{li2024learning} & CVPR'24 & \underline{65.57} & \textbf{70.20} & \underline{71.43} & \underline{60.04} & 68.01 & 52.62 & \underline{64.65} \\
        \midrule
        \textbf{DDTM (Ours)} & - & \textbf{66.45} & \underline{68.59} & \textbf{72.01} & \textbf{64.05} & \textbf{71.55} & \textbf{56.48} & \textbf{66.52} \\
        $\Delta$ w.r.t. Paraformer & - & +0.88 & $-$1.61 & +0.58 & +4.01 & +3.54 & +3.86 & +1.87 \\
        \bottomrule
    \end{tabular}}
\end{table*}

\subsection{Confidence-aware Pseudo-label Supervision}
Due to the severe resolution mismatch between low-resolution LCPs and high-resolution imagery, spatial misalignment and label noise are inevitable, particularly around object boundaries and heterogeneous regions.
To mitigate the adverse effects of noisy supervision, we introduce a confidence-aware pseudo-label supervision strategy that selectively exploits reliable training samples.
The diffusion branch first produces an initial prediction $Y_1$, which is supervised by the low-resolution labels $T$ using pixel-wise cross-entropy:
\begin{equation}
\mathcal{L}_1 = \frac{1}{HW}\sum_{i=1}^{H}\sum_{j=1}^{W}\sum_{c=1}^{C} 
t_{ij}^{(c)} \log y_{1,ij}^{(c)},
\label{eq_5}
\end{equation}
where $H$ and $W$ denote the spatial dimensions, and $C$ is the number of classes.

The Transformer branch integrates both global contextual information and refined local features, and thus requires higher-quality supervision for stable optimization.
To this end, we estimate the confidence of pseudo labels using the diffusion branch features, which are less sensitive to supervision noise due to the frozen denoising prior.
Specifically, for each training iteration, we compute a class-wise prototype feature $F_c$ by averaging diffusion features that are predicted as class $c$:
\begin{equation}
F_c = \frac{\sum_{i,j} \mathbb{I}(Y_1^{ij}=c)\,F_{ij}}{\sum_{i,j} \mathbb{I}(Y_1^{ij}=c)},
\label{eq_6}
\end{equation}
where $\mathbb{I}(\cdot)$ is the indicator function.
Each pixel-level feature is then compared with its corresponding class prototype using cosine similarity.
Pixels whose similarity exceeds a predefined threshold are regarded as high-confidence samples and form a binary confidence mask $M$.

The Transformer branch prediction $Y_2$ is supervised using the filtered pseudo labels $Y_1'$, with low-confidence regions ignored during optimization:
\begin{equation}
\mathcal{L}_2 =
\frac{\sum_{i,j} \mathbb{I}(M_{ij}=1)
\sum_{c=1}^{C} y_{1,ij}^{\prime(c)} \log y_{2,ij}^{(c)}}
{\sum_{i,j} \mathbb{I}(M_{ij}=1)}.
\label{eq_7}
\end{equation}

The final training objective combines the supervision from both branches:
\begin{equation}
\mathcal{L}_{\text{total}} = \lambda \mathcal{L}_1 + (1-\lambda)\mathcal{L}_2,
\label{eq_8}
\end{equation}
where $\lambda$ is set to 0.5 in all experiments.

\section{Experiments}
\label{sec:experiments}
\subsection{Settings}
\noindent \textbf{Datasets.}
We conduct experiments on the Chesapeake Bay land cover dataset~\cite{robinson2019large}, which is a widely used benchmark for high-resolution land cover mapping under weak supervision.
The dataset covers the Chesapeake Bay watershed and adjacent regions, and is divided into 732 non-overlapping image tiles across six U.S. states.
Following the official protocol, each state is split into 100 training images, 5 validation images, and 20 test images.
Each tile has a spatial resolution of approximately $6000\times7500$ pixels.
\begin{table}[!t]
    \centering
    \caption{Land-cover class unifying relations between the LR labels (NLCD) and HR ground truths.}
    \begin{tabularx}{\linewidth}{l X X X}
        \hline
        Name        & NLCD & CCLC & Target classes \\
        \hline
        Affiliation & USGS, USA & Chesapeake Conservancy, USA &  \\
        \hline
        Resolution  & 30 meters & 1 meter &  \\
        \hline
        \multirow{15}{*}{Class} 
        & \cellcolor{lightpink} Developed open space & Roads & \multirow{4}{*}{\colorbox{targetred}{\rule{0pt}{1.5ex}\rule{1.5ex}{1.5ex}} \ Built-up} \\
        & \cellcolor{pink} Developed low c           & Building &  \\
        & \cellcolor{red} Developed medium           & Barren &  \\
        & \cellcolor{darkred} Developed high         &  &  \\

        & \cellcolor{lightgreen1} Deciduous forest   & \multirow{5}{*}{Tree canopy} & \multirow{5}{*}{\colorbox{targetgreen}{\rule{0pt}{1.5ex}\rule{1.5ex}{1.5ex}} \ Tree canopy} \\
        & \cellcolor{lightgreen2} Evergreen forest   &  &  \\
        & \cellcolor{lightgreen3} Mixed forest       &  &  \\
        & \cellcolor{lightblue1} Woody wetland       &  &  \\
        & \cellcolor{lightgray} Barren land          &  &  \\

        & \cellcolor{lightbrown1} Shrub/Scrub        & \multirow{4}{*}{Low vegetation} & \multirow{4}{*}{\colorbox{targetyellow}{\rule{0pt}{1.5ex}\rule{1.5ex}{1.5ex}} \ Low vegetation} \\
        & \cellcolor{lightyellow1} Grassland         &  &  \\
        & \cellcolor{lightyellow2} Pasture/Har       &  &  \\
        & \cellcolor{brown1} Cultivated crops        &  &  \\

        & \cellcolor{lightblue2} Herbaceous wetlands & \multirow{2}{*}{Water} & \multirow{2}{*}{\colorbox{targetblue}{\rule{0pt}{1.5ex}\rule{1.5ex}{1.5ex}} \ Water} \\
        & \cellcolor{blue} Open water                &  &  \\
        \hline
        Note: & \multicolumn{3}{l}{USGS= United States Geological Survey;} \\ 
        \hline
    \end{tabularx}
    \label{table:1}
\end{table}

\noindent \textbf{Implementation Details.}
All methods are trained using only the low-resolution historical labels to ensure a fair comparison.
Our method DDTM is implemented in PyTorch and trained on a single NVIDIA GeForce RTX 4090 GPU using the AdamW optimizer.
During training, we randomly crop 50 patches of size $224\times224$ from each high-resolution image, with a batch size of 8.
The initial learning rate is set to 0.01.
The diffusion time step is fixed to 1000.
For confidence-aware pseudo-label supervision, the cosine similarity threshold is set to 0.9 in all experiments.
Due to the category mismatch between low-resolution and high-resolution annotations, training is performed using the 16 classes defined in the NLCD labels.
During evaluation, land cover categories are unified into four basic classes following standard practice, and performance is measured using mean Intersection over Union (mIoU).
The class unification rules are summarized in Table~\ref{table:1}.

\noindent\textbf{Baselines.}
We compare DDTM with representative weakly supervised and fully supervised segmentation models, including L2HNet~\cite{li2022breaking} and Paraformer~\cite{li2024learning}, as well as general-purpose vision architectures such as TransUNet~\cite{chen2024transunet}, ConViT~\cite{d2021convit}, CoAtNet~\cite{dai2021coatnet}, MobileViT~\cite{mehta2021mobilevit}, EfficientViT~\cite{cai2023efficientvit}, UNetFormer~\cite{wang2022unetformer}, DC-Swin~\cite{wang2022novel}, and SkipFCN~\cite{li2021change}.
All baseline methods are trained and evaluated under the same data splits and supervision settings.

\begin{figure*}[h]
    \centering
    \begin{subfigure}[b]{0.19\textwidth}
        \centering
        \includegraphics[width=\textwidth]{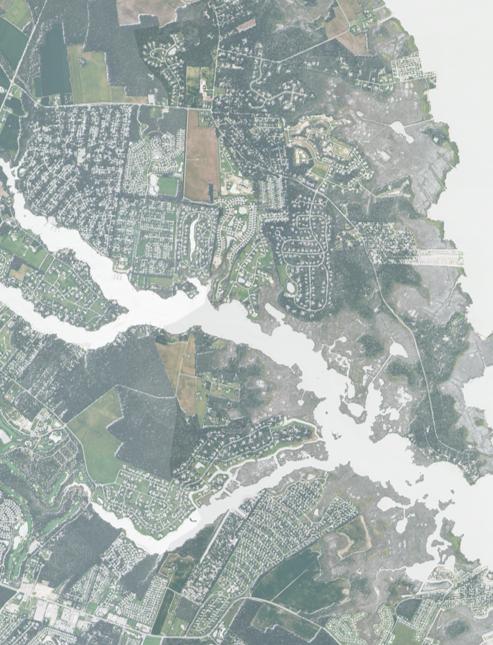} 
        \label{fig:source2} 
    \end{subfigure}
    \hfill
    \begin{subfigure}[b]{0.19\textwidth}
        \centering
        \includegraphics[width=\textwidth]{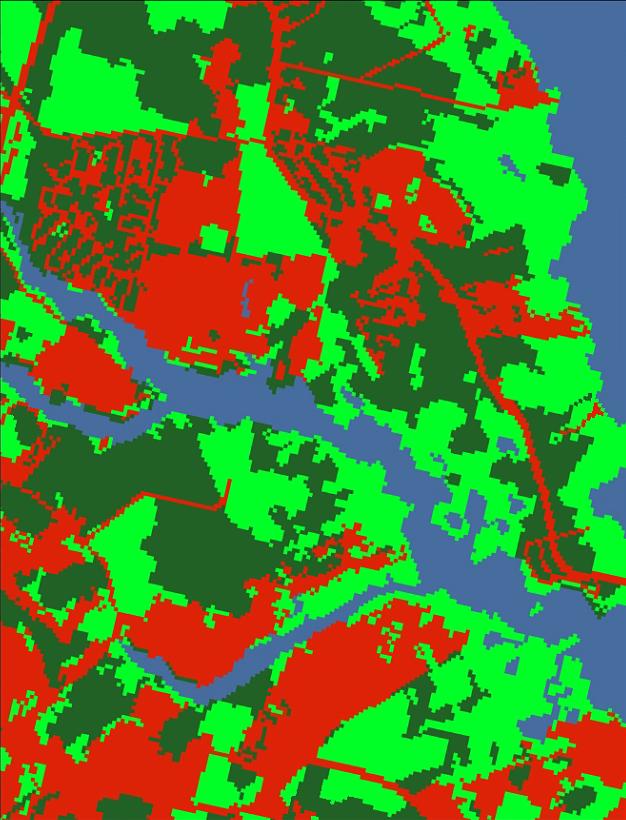} 
        \label{fig:guide2} 
    \end{subfigure}
    \hfill
    \begin{subfigure}[b]{0.19\textwidth}
        \centering
        \includegraphics[width=\textwidth]{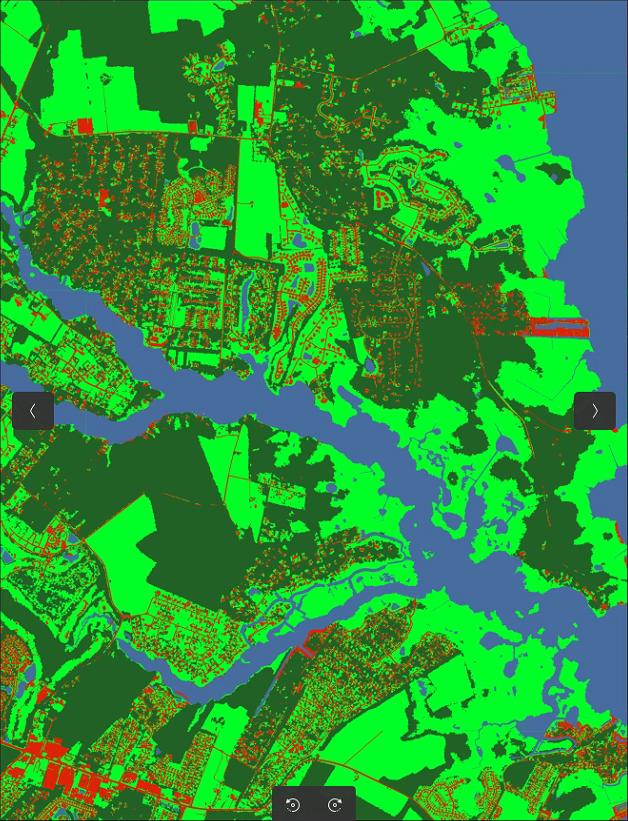} 
        \label{fig:target2} 
    \end{subfigure}
    \hfill
    \begin{subfigure}[b]{0.19\textwidth}
        \centering
        \includegraphics[width=\textwidth]{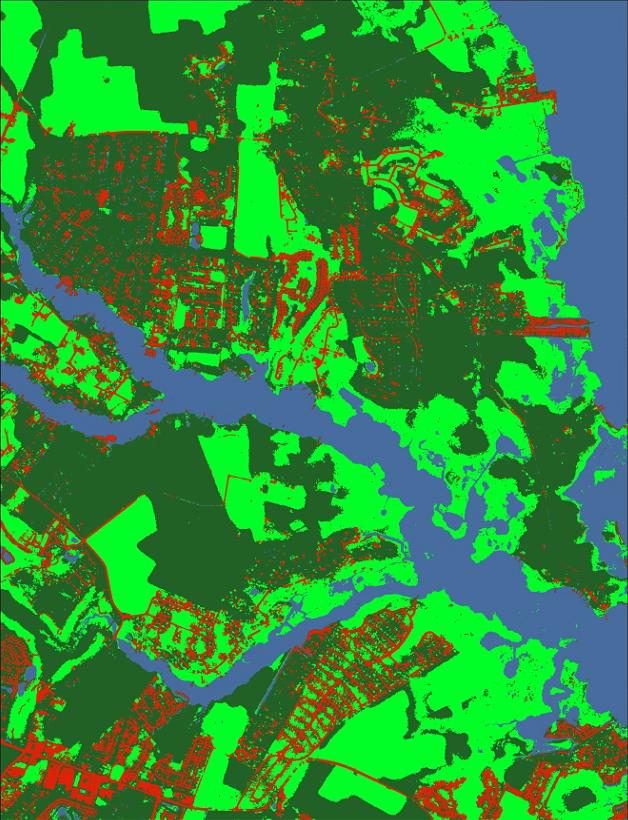} 
        \label{fig:ddtm2} 
    \end{subfigure}
    \hfill
    \begin{subfigure}[b]{0.19\textwidth}
        \centering
        \includegraphics[width=\textwidth]{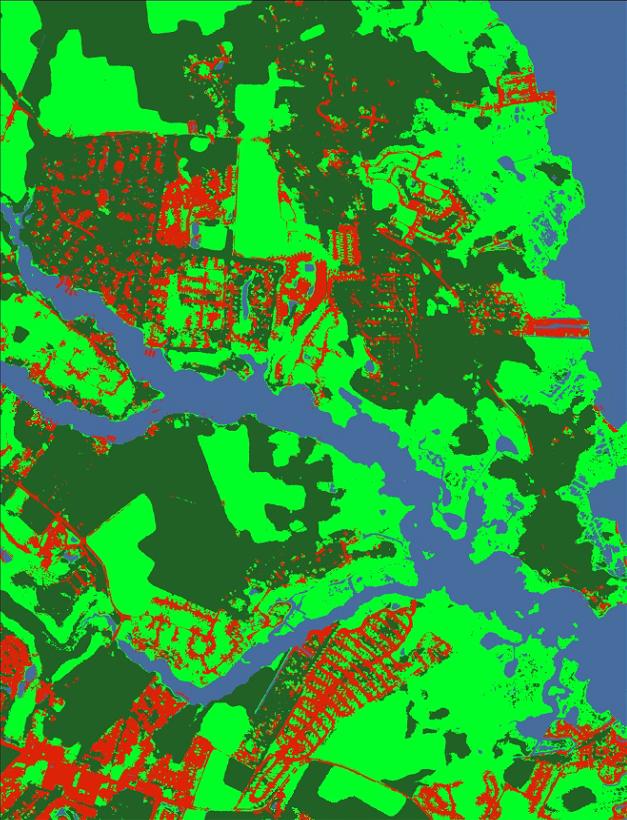} 
        \label{fig:paraformer2} 
    \end{subfigure}
     \\ 
    \begin{subfigure}[b]{0.19\textwidth}
        \centering
        \includegraphics[width=\textwidth]{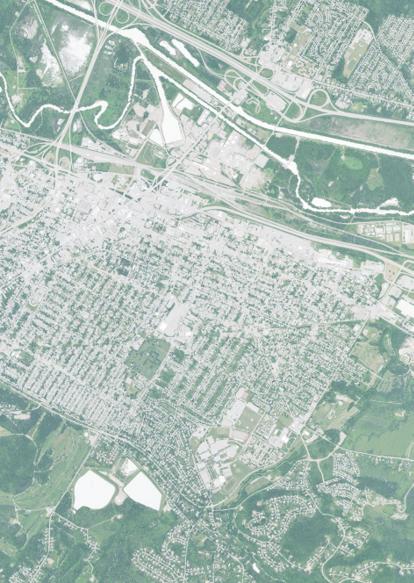}
        \caption{Source (1m/pixel)}
        \label{fig:source}
    \end{subfigure}
    \hfill
    \begin{subfigure}[b]{0.19\textwidth}
        \centering
        \includegraphics[width=\textwidth]{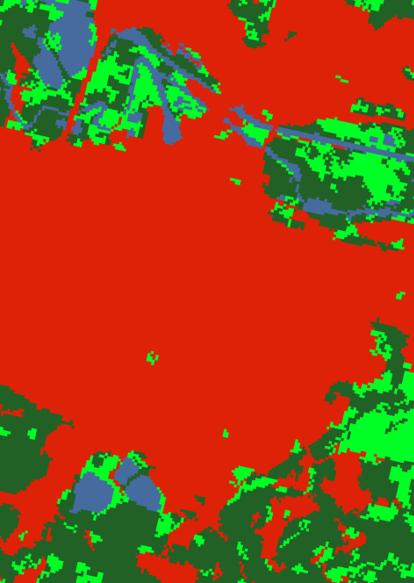}
        \caption{Guide (1m/pixel)}
        \label{fig:guide}
    \end{subfigure}
    \hfill
    \begin{subfigure}[b]{0.19\textwidth}
        \centering
        \includegraphics[width=\textwidth]{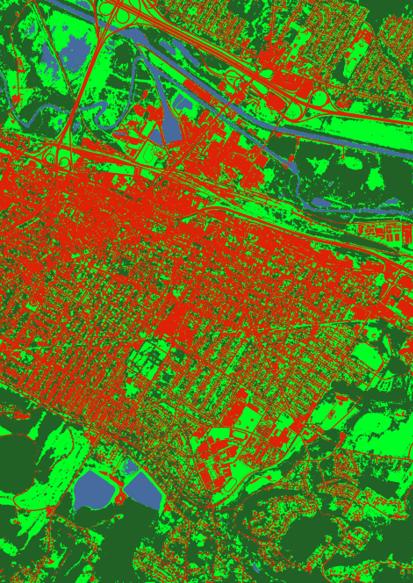}
        \caption{Target (30m/pixel)}
        \label{fig:target}
    \end{subfigure}
    \hfill
    \begin{subfigure}[b]{0.19\textwidth}
        \centering
        \includegraphics[width=\textwidth]{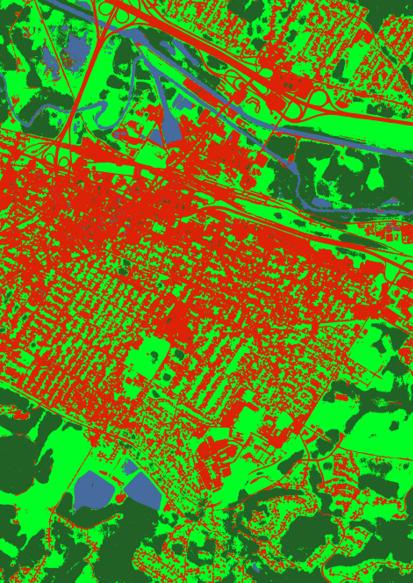}
        \caption{DDTM (ours)}
        \label{fig:ddtm}
    \end{subfigure}
    \hfill
    \begin{subfigure}[b]{0.19\textwidth}
        \centering
        \includegraphics[width=\textwidth]{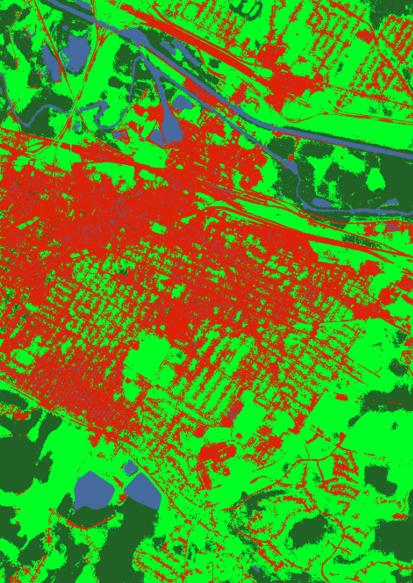}
        \caption{Paraformer}
        \label{fig:paraformer}
    \end{subfigure}
    \caption{Visualization results on the Chesapeake Bay dataset.}
    \label{fig:2}
\end{figure*}

\begin{table*}[!t]
    \centering
    \caption{Ablation results (mIoU) on the Chesapeake Bay dataset across six states.}
    \label{tab:ablation_miou}
    \setlength{\tabcolsep}{3mm}
    \resizebox{\linewidth}{!}{
    \begin{tabular}{lccccccc}
        \toprule
        Method & Delaware & New York & Maryland & Pennsylvania & Virginia & West Virginia & Average \\
        \midrule
        DDPM branch                 & 62.02 & 65.13 & 66.04 & 59.05 & 68.72 & 48.05 & 61.50 \\
        Transformer branch          & 58.76 & 65.76 & 65.32 & 54.22 & 59.96 & 50.18 & 59.03 \\
        DDTM \textit{w/o} PCEM      & 61.13 & 67.12 & 67.69 & 60.01 & 69.23 & 53.21 & 63.07 \\
        \textbf{DDTM (Full)}        & \textbf{66.45} & \textbf{68.59} & \textbf{72.01} & \textbf{64.05} & \textbf{71.55} & \textbf{56.48} & \textbf{66.52} \\
        \bottomrule
    \end{tabular}}
\end{table*}

\subsection{Experimental Results}
we summarize the quantitative and qualitative results on the Chesapeake Bay benchmark.
As shown in Table~\ref{tab:partial_miou_comparison}, DDTM achieves the best overall performance with an average mIoU of 66.52, outperforming the strongest weakly supervised baseline Paraformer by 1.87 mIoU.
Notably, DDTM consistently improves performance across five out of six states, including Delaware (+0.88), Maryland (+0.58), Pennsylvania (+4.01), Virginia (+3.54), and West Virginia (+3.86).
Paraformer performs slightly better in New York (+1.61), suggesting that the two methods exhibit complementary strengths in certain regional contexts.

Fig.~\ref{fig:2} presents qualitative comparisons between DDTM and representative baselines.
Compared with Paraformer, which tends to produce oversmoothed predictions with blurred object boundaries, DDTM generates sharper delineations and more coherent fine-scale structures.
In particular, in the overpass region shown in the upper-right area, DDTM better preserves narrow road structures and local geometric details, producing predictions that more closely align with the high-resolution ground-truth labels.
These results demonstrate the effectiveness of DDTM in jointly modeling fine-grained local semantics and long-range contextual information under cross-resolution supervision.

\subsection{Ablation Studies }
Table~\ref{tab:ablation_miou} reports ablation results evaluating the contribution of each component in DDTM.
The full model achieves the highest average mIoU of 66.52\%.
Removing either the diffusion branch or the transformer branch leads to a noticeable performance drop, indicating that local semantic refinement and global contextual modeling play complementary roles under cross-resolution supervision.
In addition, excluding the confidence-aware pseudo-label evaluation module further degrades performance, demonstrating its importance in mitigating supervision noise and stabilizing model optimization.

\section{Conclusion}
We proposed DDTM, a weakly supervised framework for cross-resolution land cover mapping that decouples local semantic refinement and global contextual modeling via a diffusion-based branch and a transformer-based branch.
To mitigate noise introduced by resolution mismatch, we further introduced a confidence-aware pseudo-label supervision module that selectively exploits reliable training signals.
Extensive experiments on the Chesapeake Bay benchmark demonstrate that DDTM achieves state-of-the-art performance under weak supervision.
In the future, we plan to improve the efficiency and scalability of the framework, and explore more lightweight backbones and more robust confidence estimation strategies for large-scale high-resolution mapping.


\bibliographystyle{IEEEbib}
\bibliography{icme2026references}

\end{document}